\documentclass[lettersize,journal]{IEEEtran}
\usepackage{amsmath,amsfonts}
\usepackage{algorithmic}
\usepackage{algorithm}
\usepackage{array}
\usepackage[caption=false,font=normalsize,labelfont=sf,textfont=sf]{subfig}
\usepackage{textcomp}
\usepackage{caption}
\usepackage{booktabs}
\usepackage{xcolor}

\usepackage{graphicx}
\usepackage{stfloats}
\usepackage{url}
\usepackage{verbatim}
\usepackage{graphicx}
\usepackage{cite}
\definecolor{cpurple}{rgb}{0.4940,0.1840,0.5560} 
\definecolor{cgreen}{rgb}{0.4660,0.6740,0.1880} 
\definecolor{corange}{rgb}{0.8500,0.3250,0.0980} 
\definecolor{cblue}{rgb}{0.0000,0.4470,0.7410}

\begin{document}

\title{Infrastructure-based Monocular 3D Vehicle Localization Framework with Experimental Validation}

\author{\'Akos T. K\"opeczi-B\'ocz, Tian Mi, G\'abor Orosz, and D\'enes Tak\'acs
\thanks{\'Akos T. K\"opeczi-B\'ocz and D\'enes Tak\'acs are with the Department of Applied Mechanics, Budapest University of Technology and Economics, Budapest, H-1111, Hungary, {\tt\small \{kopeczi,takacs\}@mm.bme.hu}}%
\thanks{Tian Mi is with the Department of Mechanical Engineering, University of Michigan, Ann Arbor, MI 48109, USA, {\tt\small tianm@umich.edu}}%
\thanks{G\'abor Orosz is with the Department of Mechanical Engineering, the Department of Civil and Environmental Engineering, and the Transportation Research Institute at the University of Michigan, Ann Arbor, MI 48109, USA.
{\tt\small orosz@umich.edu}}
}

\markboth{}%
{Shell \MakeLowercase{\textit{et al.}}: A Sample Article Using IEEEtran.cls for IEEE Journals}


\maketitle

\begin{abstract}
This paper presents a one-stage learning framework that maps monocular roadside-camera images directly to vehicle states in a ground-fixed coordinate frame. Unlike conventional approaches that first detect vehicles in the image plane and subsequently apply geometric post-processing, the proposed method leverages features from a pretrained object detector to jointly estimate each vehicle's ground-plane position, dimensions, and yaw angle. The framework therefore uses visual features not only for vehicle detection but also for direct spatial and orientation estimation. To support model training and evaluation, we develop a data-collection and label-generation pipeline based on synchronized video from a roadside camera and an unmanned aerial vehicle (UAV). Acting as a temporary top-view sensing platform, the UAV provides vehicle trajectories, dimensions, and orientations, which are transformed into the ground-fixed coordinate frame and temporally aligned with the roadside-camera images to generate ground-truth labels. The framework is evaluated using data collected during multiple experiments at the Mcity Test Facility. Results show that the proposed method can recover vehicle trajectories and orientations from monocular roadside imagery without a separate geometric post-processing stage, demonstrating its potential as a scalable approach to infrastructure-based perception at urban intersections.

\end{abstract}


\section{Introduction}

Intersections account for roughly one-quarter of traffic fatalities and approximately half of all traffic injuries in the United States \cite{FHWA_IntersectionSafety_2024}, underscoring the need for reliable situational awareness at these locations. Infrastructure-based perception is emerging as an important complement to onboard perception, particularly at urban intersections, where limited lines of sight, occlusions, and complex maneuvers create challenging traffic situations. Equipping roadside infrastructure with optical sensors enables the detection and tracking of vehicles and other road users throughout an intersection \cite{I2VReview,zhang2022roadside}. Through vehicle-to-infrastructure (V2I) communication, infrastructure-based perception systems can share information that may be unavailable to onboard sensors alone \cite{jiang2021deployment}, potentially improving both traffic safety and efficiency. Such capabilities are particularly valuable in mixed-traffic environments, where automated vehicles interact with human-driven vehicles and vulnerable road users.

Elevated roadside sensors provide broad intersection coverage and reduce inter-vehicle occlusions, making them well suited for vehicle localization. LiDAR, for example, provides accurate three-dimensional measurements that, when synchronized and calibrated with roadside cameras, can support the generation of 3D labels and their projection onto the image plane \cite{Lidar1,Lidar2,Lidar3,Lidar4,Lidar5}. However, deploying and maintaining roadside LiDAR systems remains costly, and producing large-scale labeled point-cloud datasets requires substantial annotation effort. In contrast, monocular roadside perception can build on existing camera infrastructure, offering a practical and scalable deployment path.

Monocular 3D object detection has consequently become an important research topic in vision-based perception for automated driving. Compared with stereo- or LiDAR-based detection, monocular methods require simpler hardware and have greater potential for large-scale deployment. However, they must infer depth and three-dimensional geometry from a single two-dimensional image, making their predictions particularly sensitive to occlusion, perspective ambiguity, and calibration errors.

Methods developed for vehicle-mounted monocular cameras address these challenges using multi-stage pipelines \cite{Chen_2016_CVPR,Arsalan2017,Ku2019}, geometric priors and object-interaction models \cite{Chen_2020_CVPR,Liu2020,Chabot2017}, depth-aware formulations \cite{Li_2023_AdvMono3D,Xinzhu2019,Mingyu2020}, temporal detection and tracking \cite{Hu_2019_ICCV,Time3D_2022}, and image-feature lifting into bird's-eye-view, voxel, or occupancy-like representations \cite{Rukhovich2022,Li2024,Peng2024,Li2022}. These methods, however, are primarily designed for moving, vehicle-mounted cameras and do not directly exploit the fixed calibration and ground-plane geometry available to roadside cameras.

Many existing infrastructure-based approaches use two-stage pipelines that first detect vehicles in the image plane and then estimate their ground-plane positions and orientations through geometric reasoning or post-detection processing \cite{Mahdi2023,Zwemer2022,dubska2014,sochor2018boxcars}. Although effective, these pipelines separate visual recognition from vehicle-state estimation, may not fully exploit visual features when estimating orientation, and require an additional processing stage.

Incorporating road-plane geometry directly into a one-stage detector offers a promising alternative. For example, \cite{yang2024} incorporate ground-plane information through ground-aware feature embeddings in an encoder--decoder architecture for monocular 3D object detection. Although this approach uses the ground plane as a feature-level prior, it does not explicitly parameterize the predicted vehicle states in a ground-fixed coordinate frame. A complementary strategy \cite{zhu2021} applies an image-to-ground homography to transform roadside imagery into a top-view representation, allowing vehicle localization to be formulated as one-stage oriented detection on the ground plane. However, warping the input image into a top-view representation may distort appearance information and make it more difficult to reuse features learned by detectors pretrained on perspective images. Modern one-stage detectors, including those in the YOLO family, provide transferable visual features for recognizing road users \cite{Redmon_2016_YOLO,sapkota2025ultralytics}. A key challenge is therefore to preserve and exploit these perspective-image features while directly estimating physically meaningful vehicle states in a ground-fixed coordinate frame.

In this paper, we propose an infrastructure-based monocular perception framework that maps roadside-camera images directly to vehicle states expressed in a ground-fixed coordinate frame. Starting from the one-stage YOLOv5 detector, we redesign the detection head and loss function to jointly predict each vehicle's ground-plane position, dimensions, and yaw angle in a single detection stage. This output parameterization directly constrains vehicle positions to the road surface while retaining transferable features from a detector pretrained on perspective images. Synchronized top-view observations collected by an unmanned aerial vehicle (UAV) are used to generate ground-truth vehicle states for training and evaluation; the proposed method requires only a monocular roadside camera during inference.

The main contributions of this work are as follows:
\begin{itemize}
    \item Introducing a one-stage monocular roadside-perception framework that directly predicts vehicle position, dimensions, and yaw angle in a ground-fixed coordinate frame.
    
    \item Redesigning the YOLOv5 detection head and loss function to incorporate a ground-plane output parameterization while retaining visual features learned from perspective images through transfer learning.
    
    \item Developing a synchronized UAV-roadside-camera pipeline for generating ground-truth vehicle states and aligning them with roadside-camera observations.
    
    \item Evaluating the proposed framework in both in-context and out-of-context scenarios using quantitative error measures and qualitative trajectory visualizations.
\end{itemize}

The remainder of this paper is organized as follows. Section~\ref{sec:GT} describes the experimental setup and dataset-construction pipeline, including the synchronized UAV and roadside-camera measurements, ground-truth vehicle-state extraction, camera alignment, and data filtering. Section~\ref{sec:network_training} presents the proposed ground-plane vehicle-state estimation network, including the modified detection head, loss formulation, and transfer-learning strategy. Section~\ref{sec:Results} evaluates the proposed method on the training and validation sets, as well as in-context and out-of-context test scenarios, using quantitative error measures and qualitative trajectory visualizations. Finally, Section~\ref{sec:Conclusion} summarizes the main findings, discusses the limitations of the current approach, and outlines directions for future work.

\section{Ground truth data generation}
\label{sec:GT}

This section presents the experimental setup and the generation of ground truth information. 
It provides details about the data collection process, the extraction and processing of vehicle kinematics from UAV footage, and the steps required to align the aerial and roadside perspectives for consistent dataset construction.

\subsection{Experiment description}

In a previous work \cite{Mi2025}, we developed a methodology for extracting high-precision vehicle kinematic information (position, yaw angle, velocity, and other derived quantities) directly from videos recorded by an unmanned aerial vehicle (UAV). 
The approach uses image stabilization and correlation-based template matching to identify vehicle bounding boxes and yaw angles from top-view images collected at the Mcity test facility at the University of Michigan.
The method was validated against high-precision onboard RTK GPS data, demonstrating that the UAV-based detection provided smooth trajectories and accurate kinematic state information.

To generate a training dataset for roadside camera-based vehicle detection algorithms, we conducted a new series of field measurements at the Mcity test facility. 
The data collection system consisted of two synchronized video sources. A DJI Phantom 4 drone was deployed to fly directly above the intersection, recording video at a resolution of ${1920\times1080}$ pixels that was used to generate ground truth kinematic data. 
To capture the infrastructure perspective, a roadside camera was installed on a lamppost and positioned to cover most of the intersection.

The experiments involved five vehicles across four designed experimental scenarios, each featuring specific trajectories to cover a variety of maneuvers (including straight crossings, left and right turns, following behaviors, and other vehicle interactions) to capture the complex dynamics of urban traffic. 
Each of the four scenarios was repeated five times (referred to as runs), yielding 20 distinct measurements in total.
For each run, we recorded video data simultaneously from top view and roadside perspectives, producing a dataset well-suited for training and validating our machine learning approach for infrastructure perception.

\subsection{Data collection and processing}

Our machine learning framework utilizes a high-fidelity dataset where the ground truth is derived from top-view observations. In this study, we distinguish between measured quantities (extracted directly from the top view) and derived quantities (calculated via numerical differentiation and filtering).
We define a right-handed Cartesian coordinate system $\mathcal{F}\left(x,y,z\right)$ fixed to the ground plane at the center of the intersection, where the $z$ axis points vertically upwards. We assume that the ground plane is flat. The data are collected at discrete time steps ${t_i=i\Delta t}$, where ${i=1,2,\ldots}$ and ${\Delta t \approx 17\,{\rm ms}}$ corresponding to the video frame rate (60 Hz). 
It is important to adjust the top-view footage via image stabilization to match the same frame $\mathcal{F}$. 
The primary measurements are obtained from the top-view drone footage using the template matching and stabilization workflow described in \cite{Mi2025}. 

For a vehicle at time step $t_i$, the measured state vector (ground truth data) is collected in the vector
\begin{equation}
\mathbf{m}^{\rm gt}_i = \begin{bmatrix} x_i \ y_i \ \psi_i \end{bmatrix}^\top,
\end{equation}
where $x_i$ and $y_i$ represent the coordinates of the vehicle's geometric center in the frame $\mathcal{F}$, and $\psi_i$  is the yaw angle, defined as the angle between the vehicle's longitudinal axis and the $x$ axis (pointing East). 
The velocity components ${v_{x,i}, v_{y,i}}$ and the yaw rate $\omega_{i}$ are derived as
\begin{align}
v_{x,i} = \frac{x_{i} - x_{i-1}}{\Delta t}, \quad \label{Eq:Proc_1} 
v_{y,i} = \frac{y_{i} - y_{i-1}}{\Delta t}, 
\quad 
\omega_i = \frac{\psi_{i} - \psi_{i-1}}{\Delta t}, 
\end{align}
while speed is calculated as the magnitude of the velocity vector:
\begin{equation}
v_i = \sqrt{v_{x,i}^2 + v_{y,i}^2}\,.
\label{Eq:Proc_4}
\end{equation}

\subsection{Ground truth data processing}

\begin{figure}
\centering
\includegraphics[width=8.6cm]{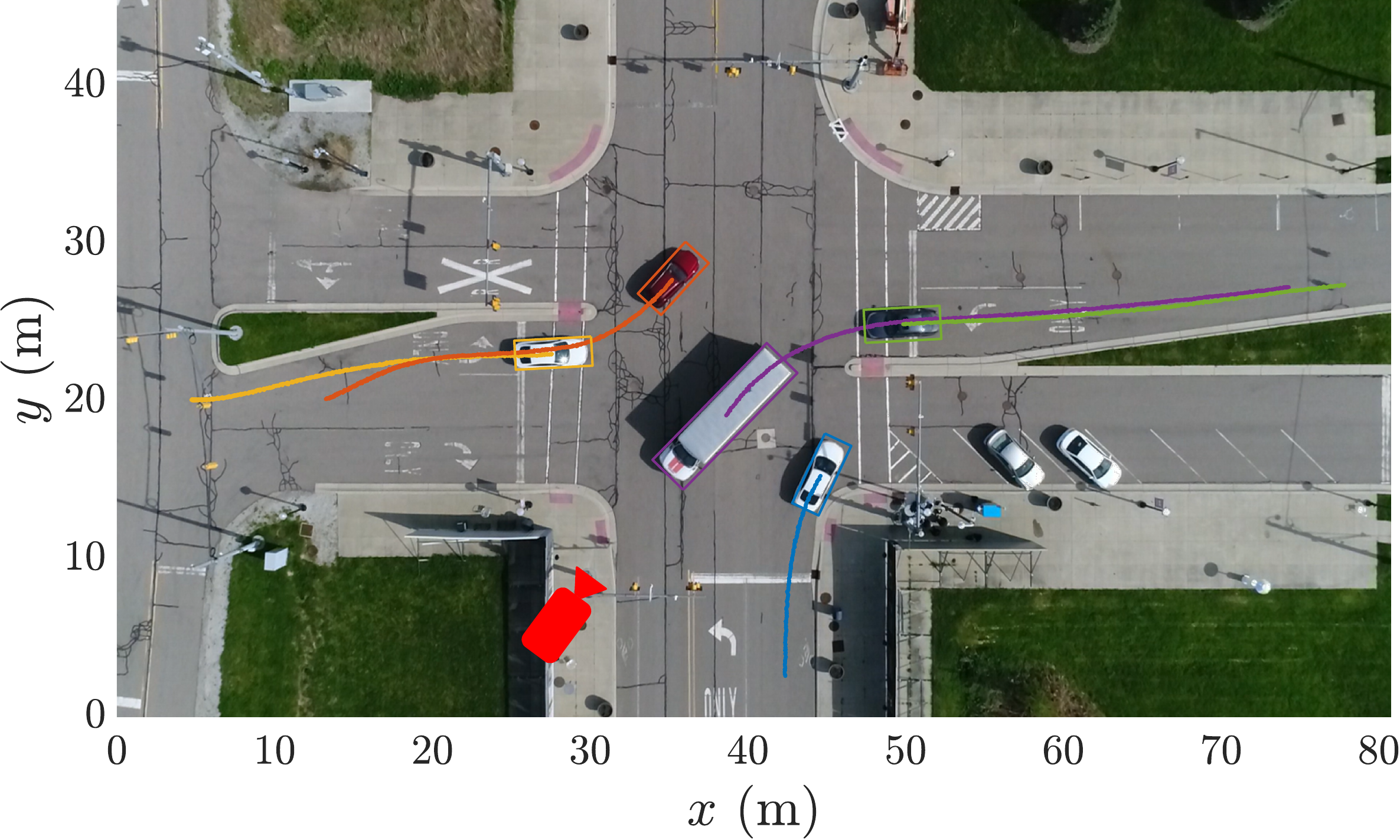}
\caption{Top view snapshot of the experimental setup at a given time instant. Colored lines indicate the trajectories of the five vehicles. The red camera icon indicates the location and orientation of the roadside camera. }
\label{Fig:GT_1}
\end{figure}

The effectiveness of the proposed machine learning framework relies heavily on the quality of the ground truth data. Consequently, the data extracted from the drone footage must be accurate and kinematically consistent to serve as a valid training target.
Fig.~\ref{Fig:GT_1} shows a top-view snapshot of an experimental run at a certain time instant and includes the past trajectories of five distinct vehicles and marks the location of the roadside camera.

\begin{figure}
\centering
\vspace{-9mm}
\includegraphics[width=0.9\linewidth]{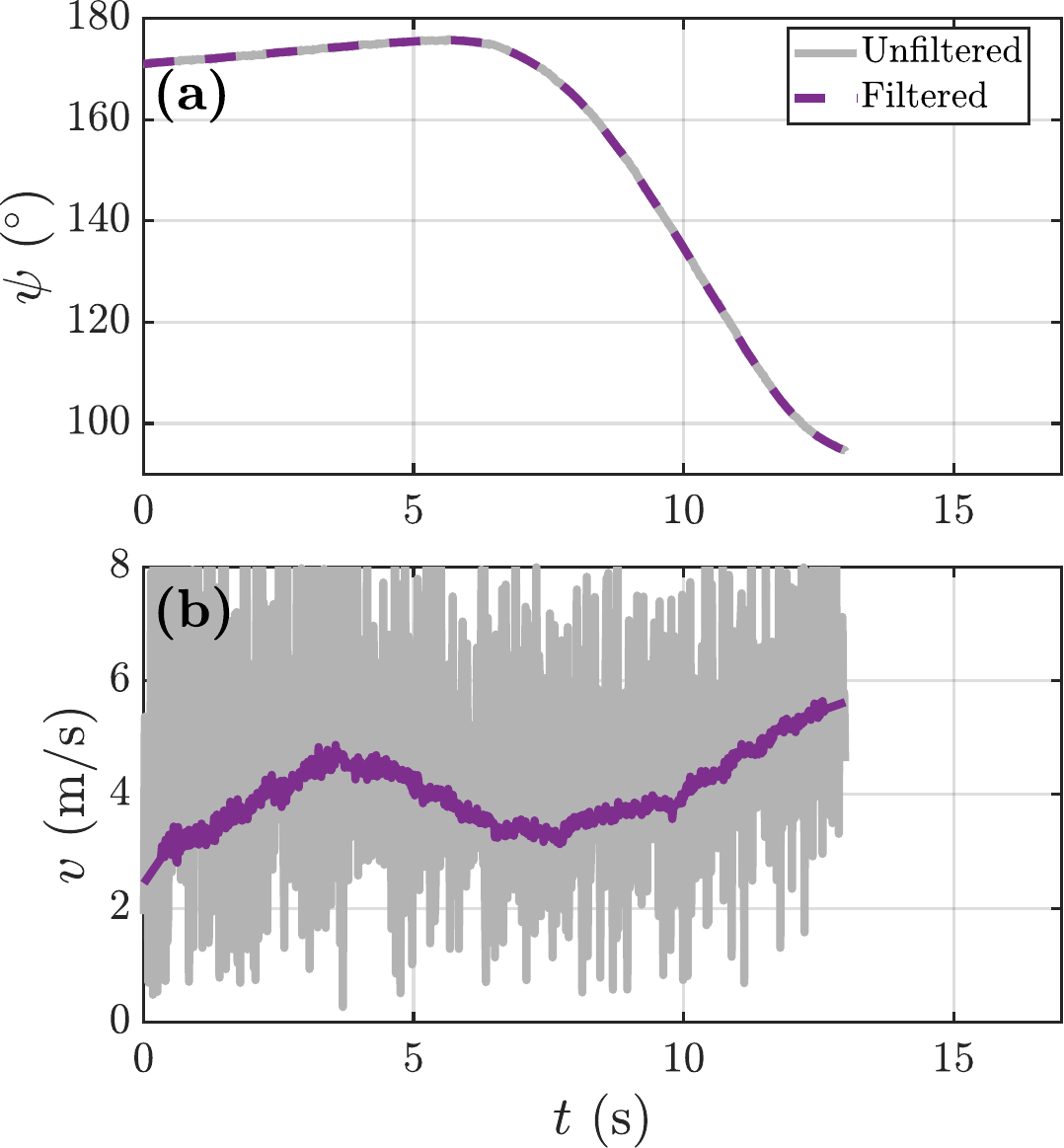}

\caption{Comparison of raw and processed ground truth data for Vehicle 1 (truck) performing a turn. 
Panel (a) and (b) show the time evolution of the yaw angle and the speed, respectively.
The gray curves represent the raw data while the purple curves show the data obtained after applying the Savitzky-Golay filter.
The noise is more significant for the speed shown in panel (b) as it is obtained via numerical differentiation of the position.}
\label{Fig:GT_2}
\end{figure}

The obtained raw position and yaw angle measurements contain measurement noise that needs to be considered when deriving kinematic data using \eqref{Eq:Proc_1}-\eqref{Eq:Proc_4}. 
Fig.~\ref{Fig:GT_2} illustrates this for Vehicle 1 (truck) by comparing the yaw angle $\psi$ and speed $v$ derived from raw measurements (gray curves) against those derived from filtered data (purple) using a Savitzky-Golay filter with polynomial order 2, window size 41 points (which is equivalent to $\approx 0.68$~s at 60~Hz sampling rate). 
This demonstrated that applying a filter is essential for presenting reliable kinematic data.

\begin{figure}
\centering 
\vspace{-15mm}
\includegraphics[width=0.9\linewidth]{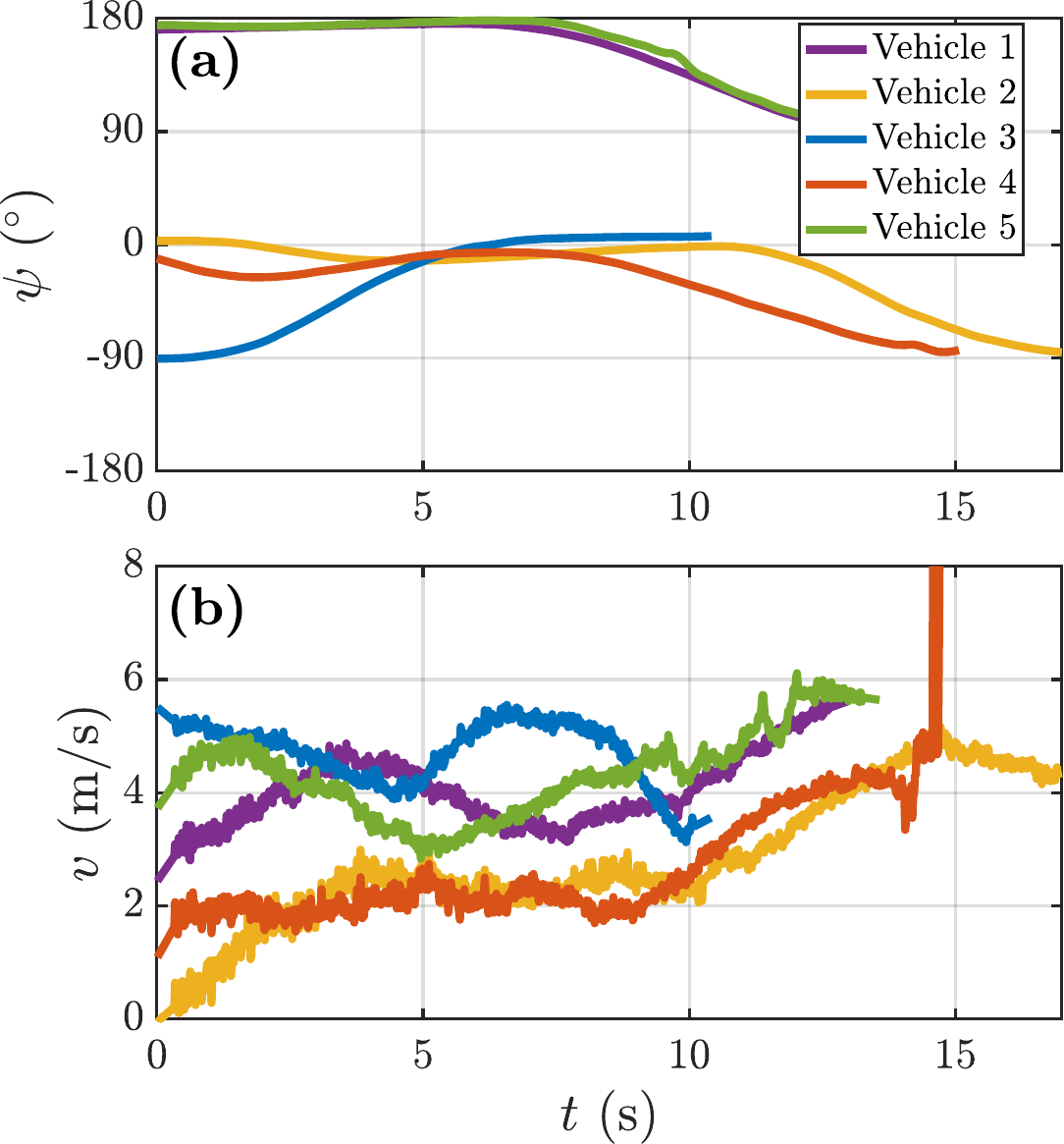}

\caption{Filtered data presented for all five vehicles. 
Panel (a) and (b) show the time evolution of the yaw angle and the speed, respectively.}
\label{Fig:GT_3}
\end{figure}

Applying this workflow to all vehicles is shown in Fig.~\ref{Fig:GT_3}, which presents the smoothed yaw angle and speed as a function of time.
This demonstrates the diversity of the collected data: some vehicles are stationary, others are accelerating, and some are executing turning maneuvers. 
This kinematic variety is essential for training, so the network captures the distinct features for these vehicles.
In Fig.~\ref{Fig:GT_3} a single run is presented for a given scenario. 
This was repeated five times, with the vehicles performing roughly the same maneuvers, and it was also repeated for the other three scenarios considered.

\subsection{Homography transformation}

To utilize the UAV-based ground truth for training our roadside view-based perception model, a mapping between the two views is required. 
This section details the processing of the data obtained from the top-view images to generate valid, view-consistent training labels for the neural network which takes the roadside-view images as input.

\begin{figure}
\centering
\includegraphics[width=3in]{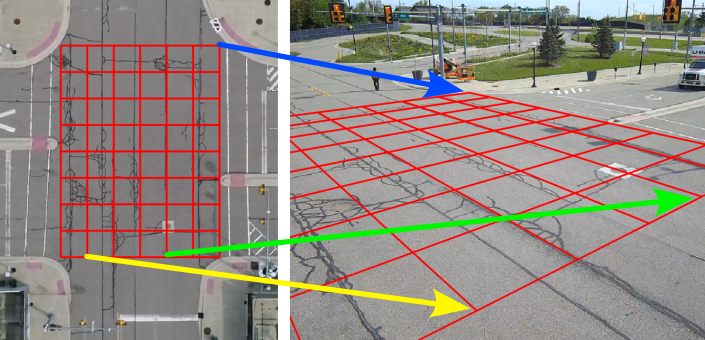}
\caption{Visualization of the homography transformation. A sample grid (red) is depicted on both the top view (left) and the roadside view (right) images. 
The arrows indicate the correspondence between specific grid points.}
\label{Fig:Hom_1}
\end{figure}

\begin{figure}
     \centering
     \includegraphics[width=\linewidth]{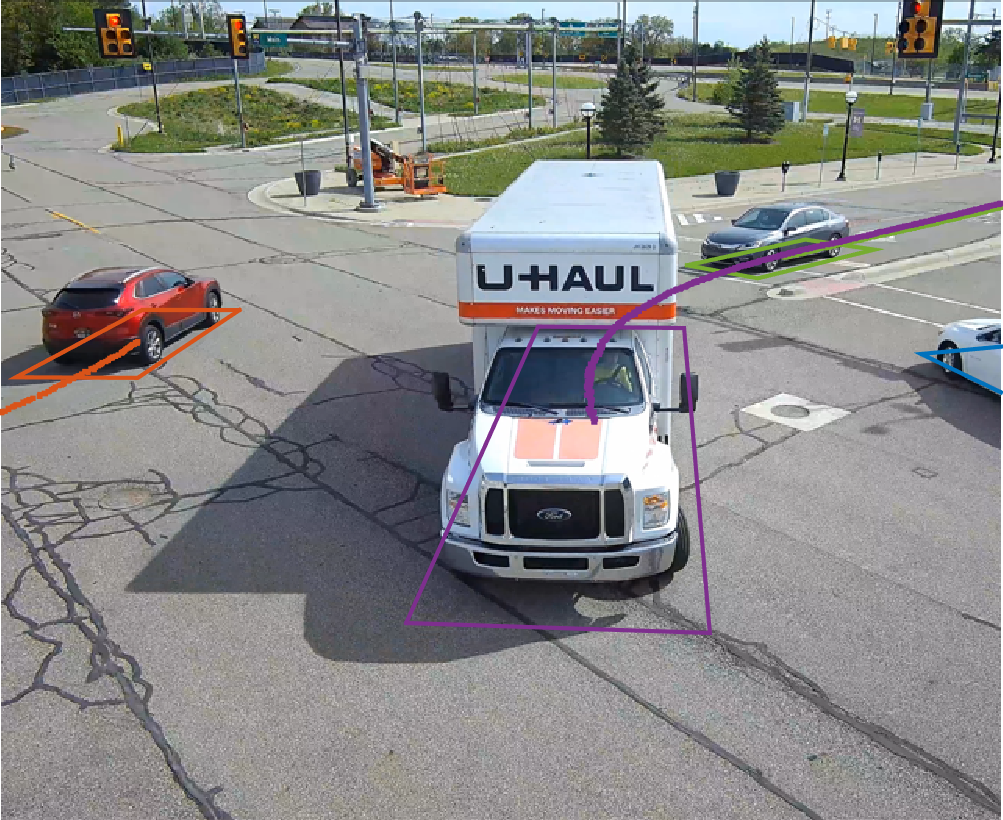}
     \caption{Roadside image taken at the time presented in Fig.~\ref{Fig:GT_1} with projected ground truth. 
     The bounding boxes and past trajectories of the vehicles are overlaid on the image. 
     }
     \label{Fig:Roadside_view}
 \end{figure}

We define the homography matrix ${\mathbf{H} \in \mathbb{R}^{3\times3}}$ that maps points from the top-view image plane to the roadside image plane. 
For a set of points 
\begin{equation}
\mathbf{P}_\mathrm{TV}=[\mathbf{x},\mathbf{y},\mathbf{1}]^\top \in \mathbb{R}^{3\times n}
\end{equation}
on the top view, the corresponding pixel location on the roadside view
\begin{equation}
\mathbf{P}_\mathrm{IMG}=[\boldsymbol{\xi},\boldsymbol{\eta},\mathbf{1}]^\top \in \mathbb{R}^{3\times n}
\end{equation}
is given by
\begin{equation}
 \mathbf{P}_\mathrm{IMG} = \mathbf{H} \, \mathbf{P}_\mathrm{TV},
 \label{Eq:Homography}
\end{equation}
Fig.~\ref{Fig:Hom_1} visualizes this mapping using a virtual grid that is depicted on both views. 
The correspondence is highlighted by arrows connecting specific grid vertices. 
The homography transformation \eqref{Eq:Homography} serves two purposes: it allows us to visualize the ground truth bounding boxes on the roadside video for verification (see Fig.~\ref{Fig:Roadside_view}), and it defines the geometric boundaries of the camera's observable area, which will be detailed in the next subsection.

Note that while we use $\mathbf{H}$ for the training dataset construction and visualization, the neural network does not receive the data mapped to the roadside view. Instead, the network implicitly learns this mapping to predict vehicle positions in frame $\mathcal{F}$ directly from the raw features on the roadside image.

\begin{figure}
\centering
\includegraphics[width=8.6cm]{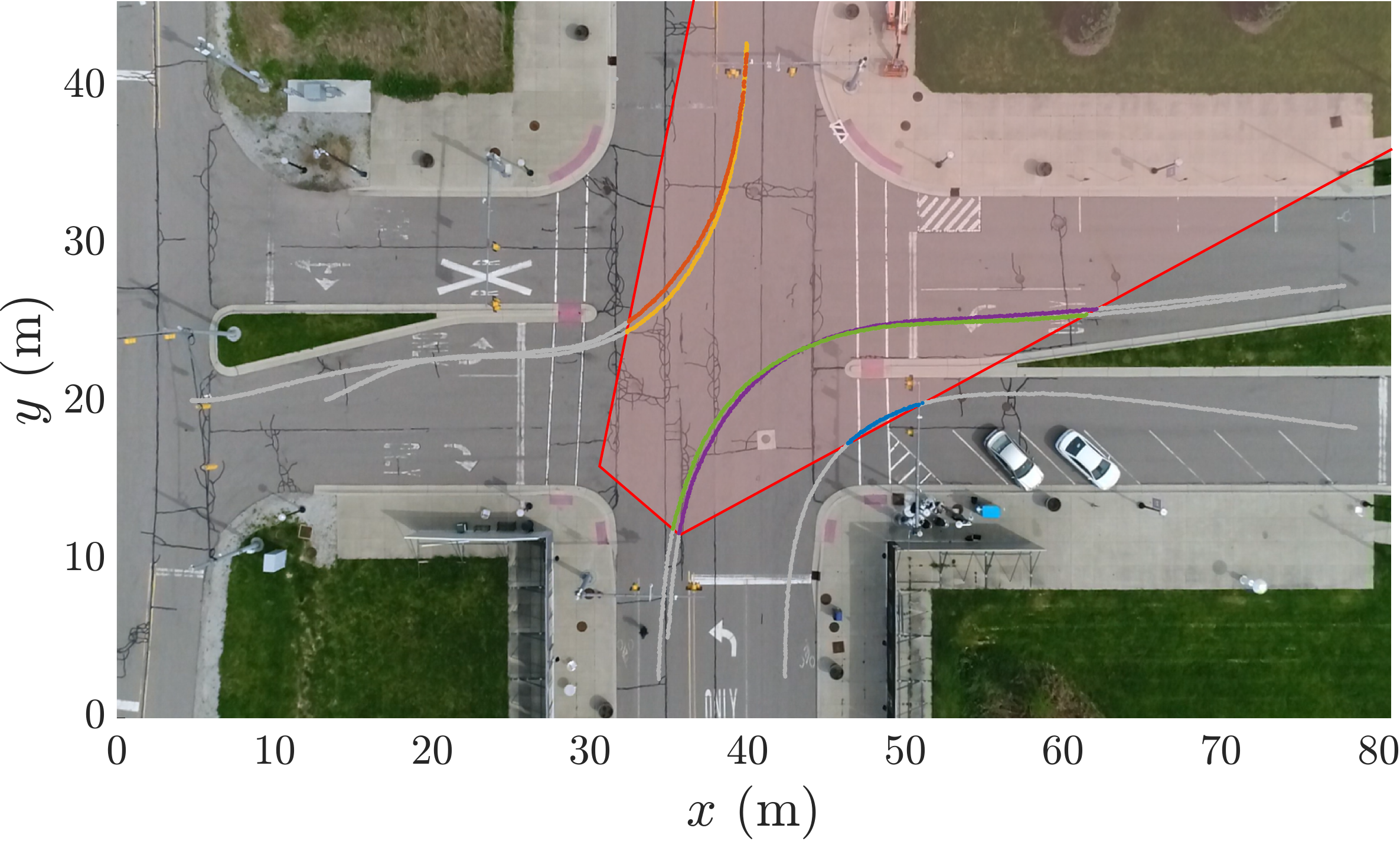}
\caption{Spatial filtering on the top view plane. The red region represents the area visible to the roadside camera. 
Vehicle trajectories are colored within the field of view and are shown in gray when outside it.}
\label{Fig:Spatial_filtering}
\end{figure}

\subsection{Spatial filtering}

A critical challenge in training the network is ensuring that the ground truth labels correspond to objects actually visible on the input image. 
The drone captures the entire intersection, but the roadside camera has a limited field of view (FoV). 
To achieve this, we project the roadside image boundaries onto the top view plane using $\mathbf{H}^{-1}$ (cf.~\eqref{Eq:Homography}), defining a polygon corresponding to the FoV of the roadside camera. Fig.~\ref{Fig:Spatial_filtering} illustrates this filtering process. 
The trajectories of all vehicles are plotted on the top view plane, with the roadside camera's FoV shaded in red. 
Colored sections correspond to data points where the vehicle's bounding box center lies within the roadside camera's FoV. 
Gray trajectories represent data points where the vehicle is outside the roadside camera's view. 
The latter are omitted during training.

Fig.~\ref{Fig:Roadside_view} illustrates the scene at the same time instant as shown in Fig.~\ref{Fig:GT_1}. 
The roadside view shows the ground-truth bounding boxes and the past trajectories projected onto the road surface. 
As shown, all four vehicles remain within the camera’s FoV, and their trajectories are visible up to the depicted time instant. 
However, as the situation evolves, Vehicle~5 (gray car) will be partially occluded by Vehicle~1 (truck). 
To solve such issues, occluded data points were excluded from the dataset, ensuring that the model is trained solely on visible features.

\section{Network topology and training}
\label{sec:network_training}

This section describes the network architecture of YOLOv5 and the modifications made to the detection heads. 
It provides details about the custom loss function formulation for the regression task and presents training losses under different layer-freezing configurations. 
We seek for the optimal freezing point that maximizes reuse of the original YOLO's pretrained features while still adapting effectively to the new task.

\subsection{Base model and modifications}

\begin{figure}
\centering
\includegraphics[scale=1]{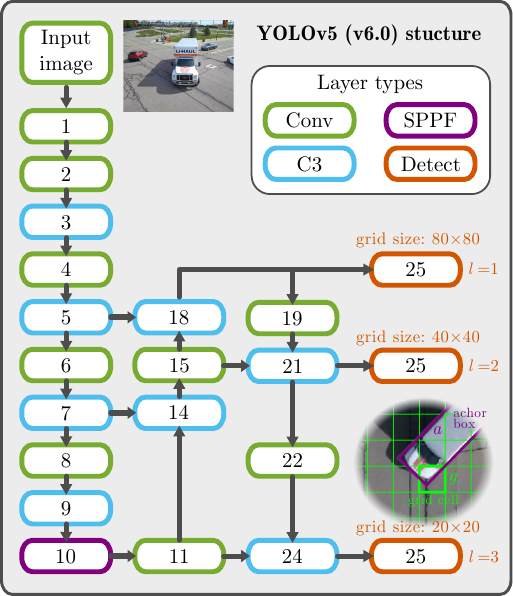}
\caption{Simplified visualization of the YOLOv5 (v6.0) structure. 
Layers with trainable weights are shown and numbered. 
Intermediate untrainable layers (like upsampling and concatenation) are not shown.
The Conv stands for the CBS module, which is composed of a standard convolution, batch normalization~\cite{batchnorm} layer and SiLU~\cite{silu} activation function. 
The C3 is a Cross Stage Partial~\cite{CSP} module with 3 convolutions. 
The SPPF\cite{SPP} stands for fast Spatial Pyramid Pooling. 
The Detect modules are custom convolutional detection heads; the corresponding output grid sizes are shown above the modules and $l$ denotes the output layer index.} 
\label{Fig:YOLOStruct}
\end{figure}

The proposed framework builds on the YOLOv5 object detection architecture, see Fig.~\ref{Fig:YOLOStruct}. 
This combines a Convolutional backbone with Spatial Pyramid Pooling (Layers~1-10) for feature extraction. The neck (Layers~11-24) consists of a Path Aggregation Network. 
The head (Layer~25) uses task‑specific detection heads. 
This network was originally designed to map a monocular image to a set of bounding boxes and class confidences defined directly on the input image plane.
Each head predicts for every spatial location on variable grid sizes. 
In each grid cell the bounding box center coordinates, width, length and class probabilities are predicted. 

In contrast, our main objective here is to utilize the constraint that the vehicle can only be placed on the ground plane. 
This is enforced by fixing the output space of the network to the base plane of the ground-fixed coordinate frame $\mathcal{F}$. In particular, we keep the backbone and the neck structure of the original network unchanged and introduce targeted modifications in the detection heads. 
In our formulation, the class predictions are reduced to a single class and the output channels are redesigned to include the orientation of the vehicle defined in the ground-fixed frame. 

Specifically, each prediction vector contains the probability $p$ of an object being present in the grid-cell, the normalized top-view position ($\hat{x}$ and $\hat{y}$), width and length ($\hat{w}$ and $\hat{l}$), and yaw angle ($\hat{\psi}$) of the bounding box:
\begin{equation}
    \mathbf{p} = 
    \begin{bmatrix}
    p & \hat{x} & \hat{y} & \hat{w} & \hat{l} & \hat{\psi}  
    \end{bmatrix}^\top \, ,
    \label{output}
\end{equation}
where hat refers to normalization that scales each quantity to ${[0, 1]}$.
A key design choice is that the network no longer parameterizes bounding boxes in the roadside image plane. Instead, the top-view output grid is fixed in the ground-fixed frame $\mathcal{F}$, and the network learns the underlying geometry and homography implicitly through its weights. 
The roadside image serves purely as input, while all targets are expressed in the  frame $\mathcal{F}$. 

The architectural changes are intentionally confined to the detection heads to maximize reuse of the original network structure. Only the final layer (Layer 25) that produces task-specific outputs is replaced to match the new kinematic output definition.
Note that our solution is not limited to the YOLOv5 network. 
Similar modifications can be applied to other object detection algorithms, particularly one-stage detectors. 
Similar developments can be observed in the YOLO family, starting with version 8, with Oriented Bounding Box detection. 
However, those solutions still do not decouple the input and output spaces, which is essential to our concept.

\subsection{Loss function formulation}

In the original YOLOv5, the bounding box regression loss is computed using intersection over union (IoU) between the predicted and ground truth boxes in the input image plane. 
For rotated bounding boxes with non‑zero yaw angles, however, exact IoU calculation becomes computationally expensive due to the need for polygon intersection algorithms. 
We therefore replace the IoU based box loss with a weighted mean squared error formulation applied directly to the predicted quantities in frame $\mathcal{F}$.

For each grid cell index $g$ and anchor index $a$ of the $l$-th output layer where a ground truth object is present as a positive sample (see Fig.~\ref{Fig:YOLOStruct}), the normalized position $\hat{x}_{l,a,g}^{\text{gt}}$, $\hat{y}_{l,a,g}^{\text{gt}}$, width/length $\hat{w}_{l,a,g}^{\text{gt}}$, $\hat{l}_{l,a,g}^{\text{gt}}$, and yaw angle $\hat{\psi}_{l,a,g}^{\text{gt}}$ are calculated from the ground truth box.
The number of detection layers is ${N_\mathrm{L}=3}$ and number of anchor boxes is ${N_\mathrm{A}=3}$. 
Let $N_\mathrm{G}( l)$ denote the total number of positive samples in layer $l$.

The total box loss is then given by the weighted sum
\begin{equation}
\mathcal{L}_{\text{box}} = \alpha \mathcal{L}_{\text{pos}} + \beta \mathcal{L}_{\text{size}} + \gamma \mathcal{L}_{\text{yaw}},
\end{equation}
where ${\alpha,\beta,\gamma}$ are tunable hyperparameters,
the position loss measures the offset error within the grid cell:
\begin{equation}
\mathcal{L}_{\text{pos}}
=
\sum_{l=1}^{N_\mathrm{L}}
\sum_{a=1}^{N_\mathrm{A}}
\sum_{g=1}^{N_\mathrm{G}( l )}
\left[
\left(\hat{x}_{l,a,g} - \hat{x}^{\text{gt}}_{l,a,g}\right)^2
+
\left(\hat{y}_{l,a,g} - \hat{y}^{\text{gt}}_{l,a,g}\right)^2
\right],
\end{equation}
the size loss penalizes deviations in bounding box dimensions:
\begin{equation}
\mathcal{L}_{\text{size}} = 
\sum_{l=1}^{N_\mathrm{L}}
\sum_{a=1}^{N_\mathrm{A}}
\sum_{g=1}^{N_\mathrm{G}( l )}
\left[
\left(\hat{w}_{l,a,g} - \hat{w}^{\text{gt}}_{l,a,g}\right)^2
+
\left(\hat{l}_{l,a,g} - \hat{l}^{\text{gt}}_{l,a,g}\right)^2
\right],
\end{equation}
and the yaw loss captures orientation mismatch:
\begin{equation}
\mathcal{L}_{\text{yaw}} = 
\sum_{l=1}^{N_\mathrm{L}}
\sum_{a=1}^{N_\mathrm{A}}
\sum_{g=1}^{N_\mathrm{G}( l)}
\left(\hat{\psi}_{l,a,g} - \hat{\psi}^{\text{gt}}_{l,a,g}\right)^2.
\label{Eq:YawLoss}
\end{equation}
Here we choose
${\alpha = 5}$, ${\beta = 1}$, ${\gamma = 10}$. 
These may be adjusted to improve accuracy, but our goal is to demonstrate the viability of the concept rather than to fine tune hyperparameters.

The overall loss combines the box loss with the unchanged objectness loss $\mathcal{L}_{\text{obj}}$, that is,
\begin{equation}
\mathcal{L} = \mathcal{L}_{\text{obj}} + \mathcal{L}_{\text{box}}.
\end{equation}
Note that classification loss is omitted as our task focuses on vehicle parameters without multi‑class categorization.

\begin{figure}[!t]
\centering
\includegraphics[width=8.2cm]{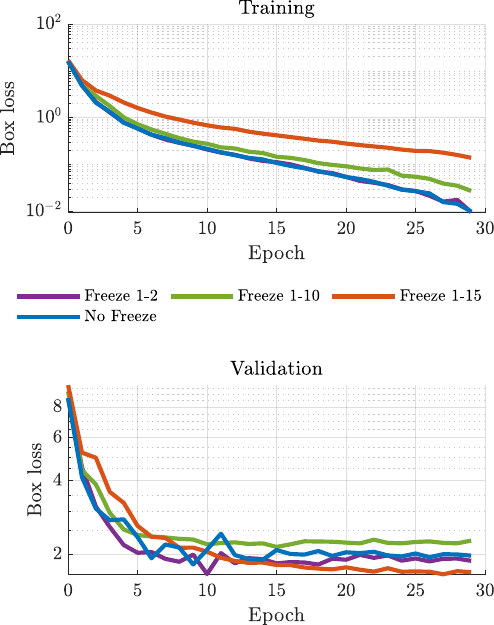}
\caption{Box losses for the training ($\mathcal{L}_{\text{box}}^{\text{train}}$) and validation ($\mathcal{L}_{\text{box}}^{\text{val}}$) for different freezing configurations.}
\label{Fig:Box_loss}
\end{figure}

\begin{figure}[!t]
\centering
\includegraphics[width=8.2cm]{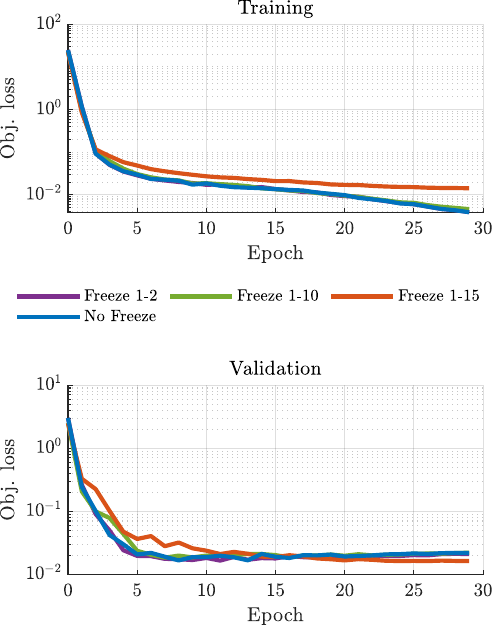}
\caption{Objectness losses for the training ($\mathcal{L}_{\text{obj}}^{\text{train}}$) and validation ($\mathcal{L}_{\text{obj}}^{\text{val}}$) for different freezing configurations.}
\label{Fig:Obj_loss}
\end{figure}

\subsection{Training results}

We adopt a stochastic gradient-based optimizer with a fixed initial learning rate and mini‑batch training. 
The backbone and neck are initialized from a pretrained YOLOv5 model (\texttt{yolov5s.pt}), while the modified detection heads are initialized with random weights. 
We train the model with different freezing configurations. That is, we freeze the weights and biases from layer 1 up to a certain layer so that the optimization algorithm treats these network parameters as fixed values. 
This reduces the number of trainable parameters and shifts the focus to learning the homography mapping and the output representation. 
In this way, we leverage the pretrained visual features for vehicle appearance while adapting only the higher-level layers to the prediction task at hand. 

\begin{table}[!t]
\centering
\begin{tabular}{@{}ll@{}}
\toprule
\textbf{Parameter} & \textbf{Value} \\
Pretrained weights & \texttt{yolov5s.pt} \\
Epochs & 30\\
Batch size & 32 \\
Image size & 640$\times$640~px\\
Optimizer & SGD \\
Learning rate & 0.01\\
Momentum & 0.937\\
Weight decay & 0.0005\\
\bottomrule
\end{tabular}
\caption{Key hyperparameters used throughout the training.}
\label{tab:training-params}
\end{table}

The base dataset contains runs 1-4 of Scenarios~1, 2, and 4. The training/validation split is 80\%-20\% derived from the random mixing of the base dataset. 
Run~5 of Scenarios~1, 2, and 4 are kept for testing purposes. These serve as in-context test sets, where the network shows its capability to predict trained positions and orientations in a known situation, but on an unseen video. 
Scenario~3 serves as the out-of-context test set to demonstrate the limitation of the network, that is, to illustrate what happens when extrapolating to untrained positions and orientations. 

The network was trained with hyperparameters listed in Table~\ref{tab:training-params}.
The training loss curves in Figs.~\ref{Fig:Box_loss} and~\ref{Fig:Obj_loss} exhibit monotonic decreases for both the box and objectness losses, demonstrating convergence across various layer freezing configurations.
The objectness term quickly saturates as the model learns to localize vehicles reliably in the top view grid, while the position and yaw components continue to improve over a longer horizon. 
The validation loss curves show a slight increase after 10-15 epochs, indicating overfitting.

Compared to the fully unfrozen network (\textcolor{cblue}{No Freeze}), the final error remains largely unchanged when freezing up to layer 10 (\textcolor{cgreen}{Freeze~1-10}), while the validation box loss saturates at distinct levels depending on the freezing point. 
Training losses are approximately one order of magnitude lower than validation losses, as highlighted by the end values in Table~\ref{tab:losses}. 
Freezing up to layers 10 and above significantly degrades performance, increasing training errors due to the exclusion of critical parameters. 
In these cases, the network struggles to adapt to the task at hand.

Notably, freezing up to layer 2 (\textcolor{cpurple}{Freeze~1-2}) yields the optimal result: 5\% reduction in validation box loss compared to the unfrozen baseline (1.8855 vs.~1.9788). 
The best validation value is achieved with freezing up to layer 15 (\textcolor{corange}{Freeze~1-15}), with an end validation loss of 1.6929. 
This, however, pairs with an order of magnitude higher training error, indicating the network's limited ability to adapt to the problem at hand. 
We investigated this case with further training, but it did not yield an improvement in the validation loss and did not achieve a better training error at the end. 
For objectness loss, shown in Fig.~\ref{Fig:Obj_loss}, the unfrozen network shows clear overfitting, with validation loss rising as training loss declines. 
Overall, similar behavior is observed: freezing up to layer 10 (\textcolor{cgreen}{Freeze~1-10}) does not affect the network's ability to adapt to the dataset, while the optimum in this range, as determined by validation, is freezing up to layer 2 (\textcolor{cpurple}{Freeze~1-2}). 

\begin{table}[!t]
\centering
\begin{tabular}{ccccc}
\hline
Freeze & $\mathcal{L}_{\text{box}}^{\text{train}}$ & $\mathcal{L}_{\text{box}}^{\text{val}}$ & $\mathcal{L}_{\text{obj}}^{\text{train}}$ & $\mathcal{L}_{\text{obj}}^{\text{val}}$ \\
\hline

\color{cblue}\textbf{No Freeze} & \color{cblue}\textbf{0.0100} & \color{cblue}\textbf{1.9788} & \color{cblue}\textbf{0.0039} & \color{cblue}\textbf{0.0217}\\
1-1 & 0.0105 & 1.9734 & 0.0039 & 0.0206\\
\color{cpurple}\textbf{1-2} & \color{cpurple}\textbf{0.0097} & \color{cpurple}\textbf{1.8855} & \color{cpurple}\textbf{0.0039} & \color{cpurple}\textbf{0.0209}\\
1-3 & 0.0098 & 1.9237 & 0.0040 & 0.0216\\
1-4 & 0.0104 & 2.1720 & 0.0039 & 0.0225\\
1-5 & 0.0100 & 2.0423 & 0.0039 & 0.0230\\
1-6 & 0.0109 & 2.3878 & 0.0040 & 0.0241\\
1-7 & 0.0113 & 3.3271 & 0.0040 & 0.0291\\
1-8 & 0.0119 & 2.9313 & 0.0040 & 0.0263\\
1-9 & 0.0141 & 2.5915 & 0.0041 & 0.0235\\
\color{cgreen}\textbf{1-10} & \color{cgreen}\textbf{0.0275} & \color{cgreen}\textbf{2.2801} & \color{cgreen}\textbf{0.0046} & \color{cgreen}\textbf{0.0220}\\
1-11 & 0.0379 & 1.9883 & 0.0050 & 0.0207\\
1-12 & 0.0379 & 1.9883 & 0.0050 & 0.0207\\
1-13 & 0.0379 & 1.9883 & 0.0050 & 0.0207\\
1-14 & 0.1013 & 1.9191 & 0.0112 & 0.0174\\
\color{corange}\textbf{1-15} & \color{corange}\textbf{0.1404} & \color{corange}\textbf{1.6929} & \color{corange}\textbf{0.0143} & \color{corange}\textbf{0.0162}\\
1-16 & 0.1404 & 1.6929 & 0.0143 & 0.0162\\
1-17 & 0.1404 & 1.6929 & 0.0143 & 0.0162\\
1-18 & 2.0802 & 3.4454 & 0.0297 & 0.0273\\
1-19 & 2.0928 & 3.4719 & 0.0301 & 0.0276\\
1-20 & 2.0928 & 3.4719 & 0.0301 & 0.0276\\
1-21 & 3.4252 & 4.4042 & 0.0310 & 0.0281\\
1-22 & 3.4276 & 4.3872 & 0.0311 & 0.0276\\
1-23 & 3.4276 & 4.3872 & 0.0311 & 0.0276\\

\hline
\end{tabular}
\caption{End loss values (at epoch 30) of the box and objectness losses for the training and validation datasets. }
\label{tab:losses}
\end{table}

Qualitative evaluation confirms that the trained network accurately recovers vehicle positions and orientations in the ground-fixed frame, even for images not used during training, supporting the effectiveness of the proposed framework.

\section{Results}
\label{sec:Results}

This section evaluates the proposed infrastructure-based perception framework for three levels of generalization. 
First, the performance on scenarios and runs included in the training and validation datasets are analyzed, establishing an upper bound on achievable accuracy under training data conditions. 
Second, the generalization to unseen runs of the same experimental scenario is examined to assess robustness to new realizations of familiar traffic configurations. 
Finally, we extrapolate to a completely unseen scenario, where vehicle trajectories and orientations are not represented in the training data, to reveal the limitations of the method. 
Throughout this section, we provide both quantitative error metrics relative to the ground truth and qualitative visualizations of trajectories in the top-view and roadside perspectives. 
In particular we depict the yaw angle error ${\Delta\psi=\psi - \psi^{\text{gt}}}$ and the speed error ${\Delta v = v - v^{\text{gt}}}$.

\subsection{Training dataset}

\begin{figure*}
\centering
\includegraphics[width=18.4 cm]{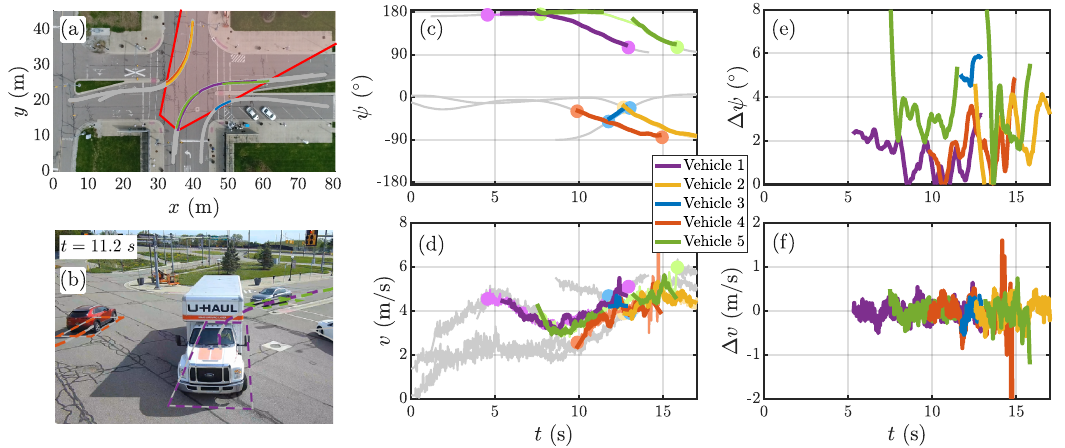}
\caption{Training and validation results on the base dataset.
(a)-(b) Ground truth trajectories are shown in gray, and network predictions are shown using dark color.
(c)-(f) Ground-truth yaw angle and speed are shown by light colors inside the field of view, while outside values are shown in gray. The network predictions are shown using dark colors.}
\label{Fig:Train_Results_Data}
\end{figure*}

We evaluate the performance of the proposed neural network by comparing its predictions against ground truth measurements. The results presented in this subsection include samples from both training and validation sets.

Figure~\ref{Fig:Train_Results_Data}(a) illustrates the comparison in the top view, where the ground truth trajectories are shown in gray, with the neural network predictions overlaid in color. 
The good alignment between the predicted and ground truth trajectories demonstrates that the model captures positions with no visible degradation between training and validation instances.
Fig.~\ref{Fig:Train_Results_Data}(b) presents a roadside view image (from the training set) at a selected time instant (corresponding to the time instant plotted in Figs.~\ref{Fig:GT_1} and \ref{Fig:Roadside_view}). 
The corresponding trajectories are depicted using dashed colored lines and the bounding boxes are also visualized. 
The close correspondence between predicted and ground truth orientations and positions indicates that the network accurately captures the data.

The time evolution of yaw angle and speed are displayed in Fig.~\ref{Fig:Train_Results_Data}(c) and (d). 
Ground truth trajectories inside the roadside camera's field of view are shown using light colors while outside they are shown in gray. 
The entry points to the field of view are highlighted by dots.
The trajectory sections predicted by the network are depicted using dark colors. 
The predicted signals closely follow the ground truth for both training and validation. 
The yaw angle error, shown in Fig.~\ref{Fig:Train_Results_Data}(e), remains predominantly under 4 degrees, while the speed error, depicted in Fig.~\ref{Fig:Train_Results_Data}(f), stays below $0.5$~m/s for most of the time. 
These results indicate that the model achieves high accuracy in both orientation and velocity estimation, with consistent performance across different data splits.

\subsection{In-context test set}

\begin{figure*}
\includegraphics[width=18.4 cm]{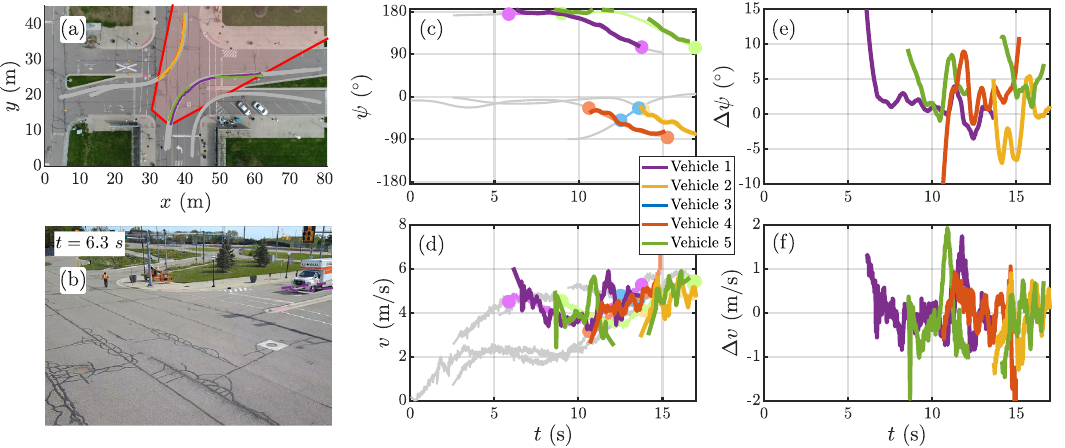}
\caption{In-context test results on a held-out run from the same experimental scenarios. 
(a)-(b) Ground truth trajectories are shown in gray, and network predictions are shown using dark color.
(c)-(f) Ground-truth yaw angle and speed are shown by light colors inside the field of view, while outside values are shown in gray. The network predictions are shown using dark colors.}
\label{Fig:In_context_Results_Data}
\end{figure*}

Here we evaluate the network’s generalization to unseen runs from the same experimental scenarios (referred to as the in-context test set). 
Specifically, we use run 5 of Scenario~1. 

Fig.~\ref{Fig:In_context_Results_Data}(a) presents the positional results in the top view.  
Although positional accuracy degrades compared to the training set -- as expected for unseen data-- the predicted trajectories closely track the ground truth ones for four vehicles. 
Notably, the blue trajectory (Vehicle~3) is not detected, as it is positioned near the edge of the roadside camera’s field of view.

The yaw angle and speed over time are shown in Fig.~\ref{Fig:In_context_Results_Data}(c) and (d), respectively. 
The network outputs generally align with the ground truth, indicating consistent performance despite unseen inputs. However, Fig.~\ref{Fig:In_context_Results_Data}(e) and (f) reveal increased errors relative to the training set.
The largest errors occur during vehicle entry into or exit from the roadside camera's field of view. 
Such an instance is depicted in Fig.~\ref{Fig:In_context_Results_Data}(b), where Vehicle~1 is entering the field of view. 
Nevertheless, the network consistently reproduces trajectories and kinematic profiles in in-context scenarios, supporting its potential for real-world deployment.

\subsection{Out-of-context test set}

\begin{figure*}
\includegraphics[width=18.4 cm]{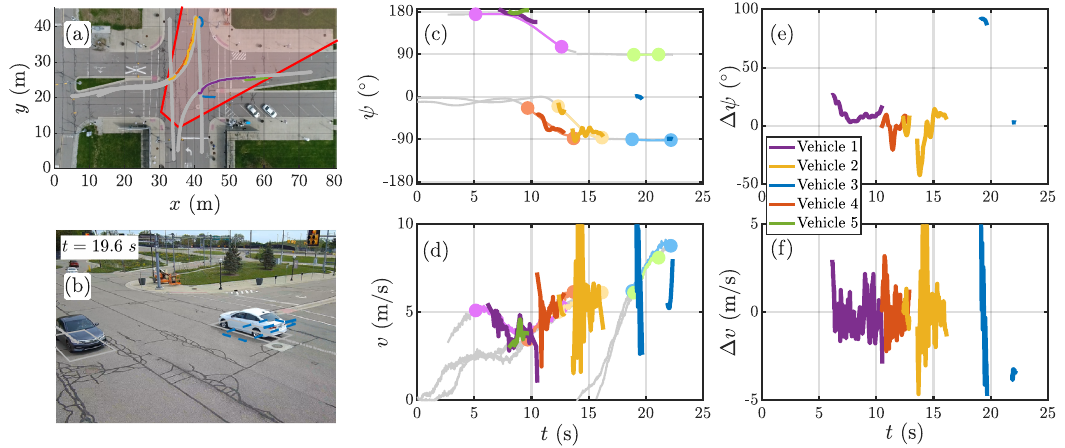}
\caption{Out-of-context test results on an unseen experiment of an unseen scenario. 
(a)-(b) Ground truth trajectories are shown in gray, and network predictions are shown using dark color.
(c)-(f) Ground-truth yaw angle and speed are shown by light colors inside the field of view, while outside values are shown in gray. The network predictions are shown using dark colors. }
\label{Fig:Out_context_Results_Data}
\end{figure*}

To assess the limits of the proposed method, we evaluate the network on an out-of-context test set (Scenario~3), which was excluded from training and validation. 
This includes trajectories and vehicle interactions that were not represented in the training data.

In Fig.~\ref{Fig:Out_context_Results_Data}(a), Vehicles~1, 2, and 4 move along trajectories that were present in the training set, whereas Vehicles~3 and 5 move along trajectories that were not part of the training distribution.
Familiar trajectory segments are reconstructed correctly (Vehicles~1, 2, and 4) even though the specific scenario has not been seen by the network. 
Unfamiliar trajectory segments are recovered only partially. 
For example, the network incorrectly associates Vehicle~5 (green) with Vehicle~1 (purple), even though Vehicle~5 is not present. 
This is because in the training data, Vehicle~5 closely followed Vehicle~1.
In addition, the trajectory of Vehicle~3 is also recovered partially. 
The second detected segment is correct, since the vehicle passed through that region along an alternative path during training.
During the first segment the vehicle is projected onto a trajectory that existed in the training data but the vehicle's orientation is perpendicular to its actual one, see Fig.~\ref{Fig:Out_context_Results_Data}(b). 

Figure~\ref{Fig:Out_context_Results_Data}(c) and (d) shows that, for correctly recovered segments, the yaw angle and speed values follow the ground truth. 
For these trajectories, the errors in Fig.~\ref{Fig:Out_context_Results_Data}(e) and (f) are in the same order of magnitude as in Fig.~\ref{Fig:In_context_Results_Data}(e) and (f). 
This indicates that, even though the training dataset is limited in diversity, the network captures the essential features well.

\section{Conclusion}
\label{sec:Conclusion}

This paper presented a monocular roadside-perception framework that maps infrastructure-camera images directly to vehicle states, including ground-plane position, dimensions, and yaw angle. By reformulating the detection task in a ground-fixed coordinate frame and training a modified YOLOv5 architecture using ground-truth labels derived from synchronized top-view UAV imagery, the proposed method learns the relationship between image features and vehicle states without requiring explicit 3D reconstruction or geometric post-processing during inference.

The results demonstrate that the proposed network can recover vehicle trajectories and orientations on both the training set and the in-context test scenarios. In particular, the model generalizes effectively to previously unseen recordings containing familiar traffic maneuvers. Yaw estimation performance, however, decreases as vehicles enter or leave the camera's field of view, where partial visibility, image-boundary truncation, and occlusion reduce the available visual information.

The out-of-context evaluation further highlights the limitations of the proposed approach under distribution shift. Although trajectory segments similar to those represented in the training data are reconstructed reliably, previously unseen trajectory configurations are recovered only partially. In some cases, the model also conflates the trajectories of vehicles whose movements were closely associated in the training data. These findings suggest that the network learns useful visual and geometric priors from the available observations but has limited ability to extrapolate to motion patterns that are not adequately represented during training.

Overall, the proposed approach offers a practical and scalable framework for extracting vehicle states and trajectory-level traffic information from monocular roadside cameras. Future work will focus on improving generalization to unseen trajectories by expanding the diversity and scale of the training data using the presented collection and labeling workflow. Additional directions include incorporating temporal information and kinematics-based motion models, developing augmentation strategies that better represent variations in trajectories and object visibility, and evaluating the framework across a broader range of intersections, camera viewpoints, and traffic conditions.

\section*{Acknowledgements}

This research was supported by the University of Michigan’s Center for Connected
and Automated Transportation through the US DOT grant 69A3552348305 and by the National Research, Development and Innovation Office of Hungary under grant
no. NKFI-146201. Akos Kopeczi-Bocz was supported by the Doctoral Excellence Fellowship Programme (DCEP) of Hungary.

\bibliographystyle{IEEEtran}
\bibliography{ref.bib}

@misc{FHWA_IntersectionSafety_2024,
  author       = {{Federal Highway Administration}},
  title        = {About Intersection Safety},
  howpublished = {US Department of Transportation, Federal Highway Administration},
  year         = {2024},
  url          = {https://highways.dot.gov/safety/intersection-safety/about}
}

@ARTICLE{I2VReview,
  author={Creß, Christian and Bing, Zhenshan and Knoll, Alois C.},
  journal={IEEE Transactions on Intelligent Transportation Systems}, 
  title={Intelligent Transportation Systems Using Roadside Infrastructure: A Literature Survey}, 
  year={2024},
  volume={25},
  number={7},
  pages={6309-6327},
  doi={10.1109/TITS.2023.3343434}}

@article{Mi2025,
   author = {Tian Mi and Dénes Takács and Henry Liu and Gábor Orosz},
   doi = {10.1080/15472450.2024.2341395},
   issn = {1547-2450},
   issue = {5},
   journal = {Journal of Intelligent Transportation Systems},
   month = {9},
   pages = {566-578},
   title = {Capturing the true bounding boxes: vehicle kinematic data extraction using unmanned aerial vehicles},
   volume = {29},
   year = {2025}
}

@InProceedings{Chen_2016_CVPR,
  author    = {Chen, Xiaozhi and Kundu, Kaustav and Zhang, Ziyu and Ma, Huimin and Fidler, Sanja and Urtasun, Raquel},
  title     = {Monocular {3D} Object Detection for Autonomous Driving},
  booktitle = {IEEE Conference on Computer Vision and Pattern Recognition (CVPR)},
  year      = {2016}
}

@InProceedings{Chen_2020_CVPR,
  author    = {Chen, Jun and Liu, Richeng and Shi, Jianping and Zhang, Lei and Xiong, Feng},
  title     = {MonoPair: Monocular {3D} Object Detection Using Pairwise Spatial Relationships},
  booktitle = {IEEE/CVF Conference on Computer Vision and Pattern Recognition (CVPR)},
  year      = {2020}
}

@InProceedings{Hu_2019_ICCV,
  title={Joint monocular 3D vehicle detection and tracking},
  author={Hu, Hou-Ning and Cai, Qi-Zhi and Wang, Dequan and Lin, Ji and Sun, Min and Krahenbuhl, Philipp and Darrell, Trevor and Yu, Fisher},
  booktitle={IEEE/CVF International Conference on Computer Vision (ICCV)},
  pages={5390--5399},
  year={2019}
}

@inproceedings{Time3D_2022,
  title={{Time3D}: End-to-end joint monocular 3d object detection and tracking for autonomous driving},
  author={Li, Peixuan and Jin, Jieyu},
  booktitle={IEEE/CVF Conference on Computer Vision and Pattern Recognition (CVPR)},
  pages={3875--3884},
  year={2022},
  organization={IEEE}
}

@inproceedings{Zwemer2022,
  title={{3D} Detection of Vehicles from 2D Images in Traffic Surveillance.},
  author={Zwemer, Matthijs H and Scholte, Dick and Wijnhoven, Rob GJ and Peter HN de With},
  booktitle={VISIGRAPP (5: VISAPP)},
  pages={97--106},
  year={2022}
}

@article{Li_2023_AdvMono3D,
  author  = {Li, Xingyuan and Liu, Jinyuan and Ma, Long and Fan, Xin and Liu, Risheng},
  title   = {Advanced Monocular {3D} Object Detection with Depth-Aware Robust Adversarial Training},
  journal = {arXiv preprint arXiv:2309.01106},
  year    = {2023}
}

@article{Redmon_2016_YOLO,
  title={You only look once: Unified, real-time object detection},
  author={Redmon, Joseph and Divvala, Santosh and Girshick, Ross and Farhadi, Ali},
  booktitle={IEEE conference on Computer Vision and Pattern Recognition (CVPR)},
  pages={779--788},
  year={2016}
}

@article{sapkota2025ultralytics,
  title={Ultralytics {YOLO} evolution: An overview of {YOLO26}, {YOLO11}, {YOLOv8} and {YOLOv5} object detectors for computer vision and pattern recognition},
  author={Sapkota, Ranjan and Karkee, Manoj},
  journal={arXiv preprint arXiv:2510.09653},
  year={2025}
}

@inproceedings{batchnorm,
  title = 	 {Batch Normalization: Accelerating Deep Network Training by Reducing Internal Covariate Shift},
  author = 	 {Ioffe, Sergey and Szegedy, Christian},
  booktitle = 	 {32nd International Conference on Machine Learning},
  pages = 	 {448--456},
  year = 	 {2015},
  editor = 	 {Bach, Francis and Blei, David},
  volume = 	 {37},
  publisher =    {PMLR}
}

@article{silu,
  title={Sigmoid-weighted linear units for neural network function approximation in reinforcement learning},
  author={Elfwing, Stefan and Uchibe, Eiji and Doya, Kenji},
  journal={Neural Networks},
  volume={107},
  pages={3--11},
  year={2018},
  publisher={Elsevier}
}

@inproceedings{CSP,
  title={CSPNet: A new backbone that can enhance learning capability of CNN},
  author={Wang, Chien-Yao and Liao, Hong-Yuan Mark and Wu, Yueh-Hua and Chen, Ping-Yang and Hsieh, Jun-Wei and Yeh, I-Hau},
  booktitle={IEEE/CVF Conference on Computer Vision and Pattern Recognition  (CVPR)},
  pages={390--391},
  year={2020}
}

@inproceedings{SPP,
  title={Spatial pyramid pooling in deep convolutional networks for visual recognition},
  author={He, Kaiming and Zhang, Xiangyu and Ren, Shaoqing and Sun, Jian},
  booktitle={European Conference on Computer Vision},
  pages={346--361},
  year={2014},
  organization={Springer}
}

@article{Lidar1,
title = {Detection and tracking of pedestrians and vehicles using roadside {LiDAR} sensors},
journal = {Transportation Research Part C},
volume = {100},
pages = {68-87},
year = {2019},
author = {Junxuan Zhao and Hao Xu and Hongchao Liu and Jianqing Wu and Yichen Zheng and Dayong Wu}
}

@article{Lidar2,
title = {Vehicle detection and tracking using low-channel roadside {LiDAR}},
journal = {Measurement},
volume = {218},
pages = {113159},
year = {2023},
author = {Ciyun Lin and Yue Wang and Bowen Gong and Hongchao Liu}
}

@article{Lidar3,
author = {Zhang, Tianya and Jin, Peter J.},
title = {Roadside {LiDAR} Vehicle Detection and Tracking Using Range and Intensity Background Subtraction},
journal = {Journal of Advanced Transportation},
volume = {2022},
number = {1},
pages = {2771085},
year = {2022}
}

@INPROCEEDINGS{Lidar4,
  author={Zimmer, Walter and Birkner, Joseph and Brucker, Marcel and Tung Nguyen, Huu and Petrovski, Stefan and Wang, Bohan and Knoll, Alois C.},
  booktitle={IEEE Intelligent Vehicles Symposium (IV)}, 
  title={InfraDet{3D}: Multi-Modal {3D} Object Detection based on Roadside Infrastructure Camera and {LiDAR} Sensors}, 
  year={2023},
  doi={10.1109/IV55152.2023.10186723}}

@Article{Lidar5,
AUTHOR = {Tihanyi, Viktor and Tettamanti, Tamás and Csonthó, Mihály and Eichberger, Arno and Ficzere, Dániel and Gangel, Kálmán and Hörmann, Leander B. and Klaffenböck, Maria A. and Knauder, Christoph and Luley, Patrick and Magosi, Zoltán Ferenc and Magyar, Gábor and Németh, Huba and Reckenzaun, Jakob and Remeli, Viktor and Rövid, András and Ruether, Matthias and Solmaz, Selim and Somogyi, Zoltán and Soós, Gábor and Szántay, Dávid and Tomaschek, Tamás Attila and Varga, Pál and Vincze, Zsolt and Wellershaus, Christoph and Szalay, Zsolt},
TITLE = {Motorway Measurement Campaign to Support {R\&D} Activities in the Field of Automated Driving Technologies},
JOURNAL = {Sensors},
VOLUME = {21},
YEAR = {2021},
NUMBER = {6},
ARTICLE-NUMBER = {2169}
}

@INPROCEEDINGS{Arsalan2017,
  author={Mousavian, Arsalan and Anguelov, Dragomir and Flynn, John and Košecká, Jana},
  booktitle={IEEE Conference on Computer Vision and Pattern Recognition (CVPR)}, 
  title={{3D} Bounding Box Estimation Using Deep Learning and Geometry}, 
  year={2017},
  volume={},
  number={},
  pages={5632-5640},
  doi={10.1109/CVPR.2017.597}}

@INPROCEEDINGS{Ku2019,
  author={Ku, Jason and Pon, Alex D. and Waslander, Steven L.},
  booktitle={IEEE/CVF Conference on Computer Vision and Pattern Recognition (CVPR)}, 
  title={Monocular {3D} Object Detection Leveraging Accurate Proposals and Shape Reconstruction}, 
  year={2019},
  volume={},
  number={},
  pages={11859-11868},
  doi={10.1109/CVPR.2019.01214}}

@INPROCEEDINGS{Liu2020,
  author={Liu, Zechen and Wu, Zizhang and Tóth, Roland},
  booktitle={IEEE/CVF Conference on Computer Vision and Pattern Recognition Workshops (CVPRW)}, 
  title={SMOKE: Single-Stage Monocular {3D} Object Detection via Keypoint Estimation}, 
  year={2020},
  volume={},
  number={},
  pages={4289-4298},
  doi={10.1109/CVPRW50498.2020.00506}}

@INPROCEEDINGS{Chabot2017,
  author={Chabot, Florian and Chaouch, Mohamed and Rabarisoa, Jaonary and Teulière, Céline and Chateau, Thierry},
  booktitle={IEEE Conference on Computer Vision and Pattern Recognition (CVPR)}, 
  title={Deep MANTA: A Coarse-to-Fine Many-Task Network for Joint {2D} and {3D} Vehicle Analysis from Monocular Image}, 
  year={2017},
  volume={},
  number={},
  pages={1827-1836},
  doi={10.1109/CVPR.2017.198}}

@INPROCEEDINGS{Xinzhu2019,
  author={Ma, Xinzhu and Wang, Zhihui and Li, Haojie and Zhang, Pengbo and Ouyang, Wanli and Fan, Xin},
  booktitle={IEEE/CVF International Conference on Computer Vision (ICCV)}, 
  title={Accurate Monocular {3D} Object Detection via Color-Embedded {3D} Reconstruction for Autonomous Driving}, 
  year={2019},
  volume={},
  number={},
  pages={6850-6859},
  doi={10.1109/ICCV.2019.00695}}

@INPROCEEDINGS{Mingyu2020,
  author={Ding, Mingyu and Huo, Yuqi and Yi, Hongwei and Wang, Zhe and Shi, Jianping and Lu, Zhiwu and Luo, Ping},
  booktitle={IEEE/CVF Conference on Computer Vision and Pattern Recognition (CVPR)}, 
  title={Learning Depth-Guided Convolutions for Monocular {3D} Object Detection}, 
  year={2020},
  volume={},
  number={},
  pages={11669-11678},
  doi={10.1109/CVPR42600.2020.01169}}

@INPROCEEDINGS{Rukhovich2022,
  author={Rukhovich, Danila and Vorontsova, Anna and Konushin, Anton},
  booktitle={IEEE/CVF Winter Conference on Applications of Computer Vision (WACV)}, 
  title={ImVoxelNet: Image to Voxels Projection for Monocular and Multi-View General-Purpose {3D} Object Detection}, 
  year={2022},
  volume={},
  number={},
  pages={1265-1274},
  doi={10.1109/WACV51458.2022.00133}}

@ARTICLE{Li2024,
  author={Li, Yangguang and Huang, Bin and Chen, Zeren and Cui, Yufeng and Liang, Feng and Shen, Mingzhu and Liu, Fenggang and Xie, Enze and Sheng, Lu and Ouyang, Wanli and Shao, Jing},
  journal={IEEE Transactions on Pattern Analysis and Machine Intelligence}, 
  title={Fast-BEV: A Fast and Strong Bird’s-Eye View Perception Baseline}, 
  year={2024},
  volume={46},
  number={12},
  pages={8665-8679},
  doi={10.1109/TPAMI.2024.3414835}}

@INPROCEEDINGS{Peng2024,
  author={Peng, Liang and Xu, Junkai and Cheng, Haoran and Yang, Zheng and Wu, Xiaopei and Qian, Wei and Wang, Wenxiao and Wu, Boxi and Cai, Deng},
  booktitle={IEEE/CVF Conference on Computer Vision and Pattern Recognition (CVPR)}, 
  title={Learning Occupancy for Monocular {3D} Object Detection}, 
  year={2024},
  volume={},
  number={},
  pages={10281-10292},
  doi={10.1109/CVPR52733.2024.00979}}

@inproceedings{Li2022,
 author = {Li, Yanwei and Chen, Yilun and Qi, Xiaojuan and Li, Zeming and Sun, Jian and Jia, Jiaya},
 booktitle = {Advances in Neural Information Processing Systems},
 doi = {10.52202/068431-1340},
 editor = {S. Koyejo and S. Mohamed and A. Agarwal and D. Belgrave and K. Cho and A. Oh},
 pages = {18442--18455},
 publisher = {Curran Associates, Inc.},
 title = {Unifying Voxel-based Representation with Transformer for {3D} Object Detection},
 volume = {35},
 year = {2022}
}

@article{Mahdi2023,
title = {{3D-Net}: Monocular {3D} object recognition for traffic monitoring},
journal = {Expert Systems with Applications},
volume = {227},
pages = {120253},
year = {2023},
author = {Mahdi Rezaei and Mohsen Azarmi and Farzam Mohammad Pour Mir}
}

@inproceedings{dubska2014,
  title={Automatic camera calibration for traffic understanding},
  author={Dubsk{\'a}, Mark{\'e}ta and Herout, Adam and Sochor, Jakub},
  booktitle={BMVC},
  volume={4},
  number={6},
  pages={1-12},
  year={2014}
}

@article{sochor2018boxcars,
  title={Boxcars: Improving fine-grained recognition of vehicles using {3-D} bounding boxes in traffic surveillance},
  author={Sochor, Jakub and {\v{S}}pa{\v{n}}hel, Jakub and Herout, Adam},
  journal={IEEE Transactions on Intelligent Transportation Systems},
  volume={20},
  number={1},
  pages={97--108},
  year={2018},
  publisher={IEEE}
}

@article{yang2024,
  title={MonoGAE: Roadside monocular {3D} object detection with ground-aware embeddings},
  author={Yang, Lei and Zhang, Xinyu and Yu, Jiaxin and Li, Jun and Zhao, Tong and Wang, Li and Huang, Yi and Zhang, Chuang and Wang, Hong and Li, Yiming},
  journal={IEEE Transactions on Intelligent Transportation Systems},
  volume={25},
  number={11},
  pages={17587--17601},
  year={2024},
  publisher={IEEE}
}

@inproceedings{zhu2021,
  title={Monocular {3D} vehicle detection using uncalibrated traffic cameras through homography},
  author={Zhu, Minghan and Zhang, Songan and Zhong, Yuanxin and Lu, Pingping and Peng, Huei and Lenneman, John},
  booktitle={IEEE/RSJ International Conference on Intelligent Robots and Systems (IROS)},
  pages={3814--3821},
  year={2021},
  organization={IEEE}
}

@article{jiang2021deployment,
  title={On the deployment of V2X roadside units for traffic prediction},
  author={Jiang, Lejun and Moln{\'a}r, Tam{\'a}s G and Orosz, G{\'a}bor},
  journal={Transportation Research Part C},
  volume={129},
  pages={103238},
  year={2021},
  publisher={Elsevier}
}

@article{zhang2022roadside,
author = {Zhang, Rusheng and Zou, Zhengxia and Shen, Shengyin and Liu, X., Henry },
title ={Design, Implementation, and Evaluation of a Roadside Cooperative Perception System},
journal = {Transportation Research Record},
volume = {2676},
number = {11},
pages = {273-284},
year = {2022},
doi = {10.1177/03611981221092402},
URL = {https://doi.org/10.1177/03611981221092402},
eprint = {https://doi.org/10.1177/03611981221092402}
}

\vfill

\end{document}